\documentclass[a4paper,twoside]{article}

\usepackage{tabularx}
\usepackage{epsfig}
\usepackage{subcaption}
\usepackage{calc}
\usepackage{amssymb}
\usepackage{amstext}
\usepackage{amsmath}
\usepackage{amsthm}
\usepackage{xcolor}
\RequirePackage{xcolor, colortbl}
\usepackage{multicol}
\usepackage{graphicx}
\usepackage{pslatex}
\usepackage{apalike}
\usepackage{algorithm2e}
\usepackage[bottom]{footmisc}
\usepackage{SCITEPRESS}     
\usepackage{epstopdf}
\usepackage{graphicx}

\begin{document}

\title{A Novel Vision Transformer for Camera-LiDAR Fusion based Traffic Object Segmentation}

\author{
\authorname{
Toomas Tahves\sup{1}\orcidAuthor{0009-0008-0050-2146},
Junyi Gu\sup{1}\sup{2}\orcidAuthor{0000-0002-5976-6698},
Mauro Bellone\sup{3}\orcidAuthor{0000-0003-3692-0688},
Raivo Sell\sup{1}\orcidAuthor{0000-0003-1409-0206}
}
\affiliation{\sup{1}Department of Mechanical and Industrial Engineering, Tallinn University of Technology, Estonia}
\affiliation{\sup{2} Dept. of Computer Science and Engineering, Chalmers University of Technology and University of Gothenburg, Sweden}
\affiliation{\sup{3}FinEst Centre for Smart Cities, Tallinn University of Technology, Estonia}
}

\abstract{
This paper presents Camera-LiDAR Fusion Transformer (CLFT) models for traffic object segmentation, which leverage the fusion of camera and LiDAR data using vision transformers.
Building on the methodology of visual transformers that exploit the self-attention mechanism, we extend segmentation capabilities with additional classification options to a diverse class of objects including cyclists, traffic signs, and pedestrians across diverse weather conditions.
Despite good performance, the models face challenges under adverse conditions which underscores the need for further optimization to enhance performance in darkness and rain.
In summary, the CLFT models offer a compelling solution for autonomous driving perception, advancing the state-of-the-art in multimodal fusion and object segmentation, with ongoing efforts required to address existing limitations and fully harness their potential in practical deployments.
}

\onecolumn \maketitle \normalsize \setcounter{footnote}{0} \vfill

\section{\uppercase{Introduction}}
\label{sec:introduction}

This work extends our previous work on camera-LiDAR fusion transformer (CLFT)~\cite{clft}, which utilizes the encoder-decoder structure of a transformer network but uses a novel progressive-assemble strategy of vision transformers.
We elaborate on the CLFT methodology and extend segmentation with additional classification options.
Our goal is to outperform existing CNN and visual transformer models by leveraging camera and LiDAR data fusion.

Transformers ~\cite{transformers}, initially introduced for language models, rely on a mechanism called self-attention to process input data patches. 
This allows models to globally weigh the importance of different parts of input data simultaneously, thus improving computation efficiency.
Since transformers do not contain information about the order of input tokens, positional encodings are added to input embeddings to retain information which is crucial to remember in tasks such as language translation and image recognition.

Vision transformers (ViT) ~\cite{vit} apply the transformer architecture to image data by dividing images into patches and treating each patch as a token which allows models to capture global context and relationships between different parts of an image.
Dense prediction transformers (DPT) ~\cite{dpt} process image patches similarly to ViTs but focus on generating pixel-level predictions by leveraging the strengths of transformers in capturing long-range dependencies and contextual information.
Our hypothesis is that the combination of ViT and DPT can grab dependencies in the data improving the interpretation of less-represented classes in consideration that autonomous driving datasets are strongly unbalanced to vehicles. 

Following this line of research, our work provides the following main contributions:
\begin{itemize}
    \item We enhanced the CLFT model to handle a broader spectrum of traffic objects, including cyclists, signs, and pedestrians.
    \item Through extensive testing, we demonstrated that our model achieves superior accuracy and performance metrics compared to other visual transformer models.
    \item By leveraging the strengths of multi-modal sensor fusion and the multi-attention mechanism, the CLFT model proves to be a solution for diverse environmental conditions, including challenging weather scenarios.
\end{itemize}

\section{\uppercase{Related work}}
The fusion of camera and LiDAR data is a widely researched topic in multimodal fusion with applications in object detection and segmentation.
Various techniques have been proposed over the years to solve these problems, ~\cite{categories} proposed the following categorization options: signal-level, feature-level, result-level, and multi-level fusion.
Signal-level fusion depends on raw sensor data, while it is suitable for depth completion ~\cite{related-depth-completion-1} ~\cite{related-depth-completion-2} and landmark detection ~\cite{related-road-detection-1} ~\cite{related-road-detection-2}, it still suffers from loss of texture information.
Voxel grid or 2D projection are used to represent LiDAR data as feature maps, for instance, the implementation of VoxelNet ~\cite{related-feature-fusion-1} uses raw point clouds as voxels before fusing LiDAR data with camera pixels.
Result-level fusion increases accuracy by merging prediction results from different model outputs ~\cite{related-result-fusion-1} ~\cite{related-result-fusion-2}.
Through reviewing the literature, it is possible to observe that the recent trend is to shift towards multi-level fusion, which represents a combination of all other fusion strategies. 
The computational complexity resulting from LiDAR 3D data is tackled by reducing the dimensionality to a two-dimensional image to exploit the existing image processing methods. 
Our work uses a transformer-based network for integrating camera and LiDAR data in a cross-fusion strategy in the decoder layers. 

The attention mechanism introduced in the transformer architecture in ~\cite{transformers} has a tremendous impact in various fields, especially in natural language processing ~\cite{nlp-transformers} and computer vision.
One notable variant is the vision transformer (ViT) ~\cite{vit}, which excels in autonomous driving tasks by handling global contexts and long-range dependencies.
Perceiving the surrounding area in a two-dimensional plane primarily involves extracting information from camera images with notable works like bird eye view transformers for road surface segmentation presented in ~\cite{2d-paper-1}.
Other recent approaches include lightweight transformers for lane shape prediction and combined semantic and instance segmentation ~\cite{2d-paper-2}.
Three-dimensional autonomous driving perception is an extensively researched topic focusing on object detection and segmentation.
In ~\cite{3d-paper-1} DETR3D, the authors present a multi-camera object detection method, unlike others that rely on monocular images, it extracts 2D features from images and uses 3D object queries to link features to 3D positions via camera transformation matrices.
FUTR3D ~\cite{3d-paper-2} employs a query-based Modality-Agnostic Feature Sampler (MAFS), together with a transformer decoder with a set-to-set loss for 3D detection, thus avoiding using late fusion heuristics and post-processing tricks.
BEVFormer ~\cite{3d-paper-3} improves object detection and map segmentation with spatial and temporal attention layers via spatiotemporal transformers.

Recent works emphasize the fusion of camera and LiDAR data for enhanced perception.
CLFT models, for instance, process LiDAR point clouds as image views to achieve 2D semantic segmentation, bridging gaps in multi-modal semantic object segmentation.

\section{\uppercase{Methodology}}

\begin{figure*}[ht]
    \centering
    \includegraphics[width=\textwidth ]{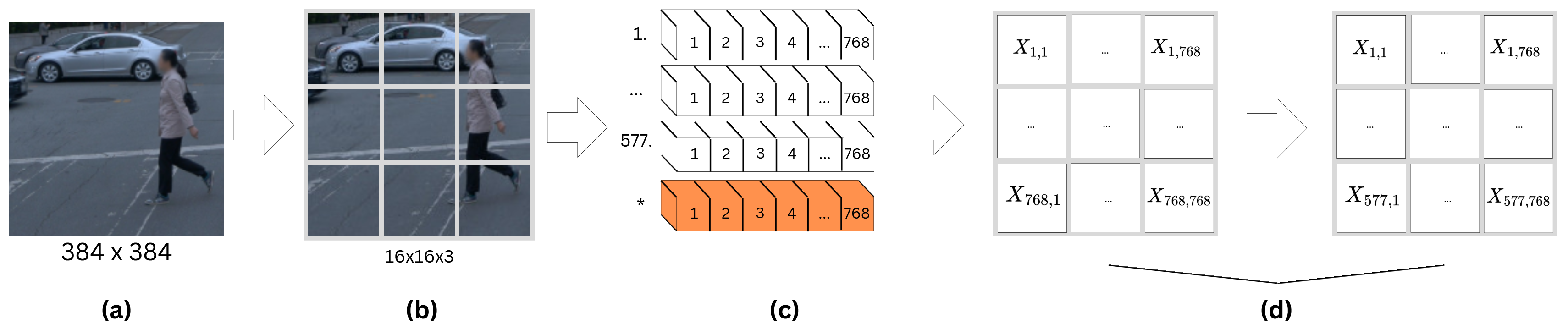}
    \caption{Embedding process for camera and LiDAR data. (a) The original image is resized to a resolution of $384 \times 384$ to standardize the input dimensions. (b) The input image is segmented into non-overlapping fixed-size patches of $16 \times 16$ pixels. (c) Patches are flattened into one-dimensional embedded vectors, with an additional positional embedding (colored in orange) added to provide spatial information. (d) The combined patch embeddings are processed through Multilayer Perceptrons (MLPs) with dimensions $E = \Bar{D} \times D$, resulting in a matrix that serves as the input for the transformer encoder. The whole figure is based on the CLFT-Base variant.}
    \label{fig:embedding}
\end{figure*}

In this section, we elaborate on the detailed structure of the CLFT network in the sequential order of data processing, aiming to provide an exclusive insight into how the sensory data flows in the network,
thus, benefits the understanding and reproducibility of our work.  

The CLFT network achieves the camera-LiDAR fusion by progressively assembling features from each modality first and then conducting the cross-fusion at the end. 
Figuratively, the CLFT network has two directions to process the input camera and LiDAR data in parallel; the integration of two modalities happens at the `fusion' stage in the network's decoder block. 
In general, there are three steps in the entire process. 
The first step is pre-processing the input, which embeds the image-like data to the learnable transformer tokens; 
the second step closely follows the protocols of ViT \cite{vit} encoders to encode the embedded tokens; 
the last step is the post-processing of the data, which progressively assembles and fuses the feature representations to acquire segmentation predictions. 
The details of the three steps are described in the following three subsections. 

\subsection{Embedding}
The camera and LiDAR input data pre-processing is independent and in parallel. 
As mentioned in Section~\ref{sec:introduction}, we select the LiDAR processing strategy to project the point cloud data onto the camera plane, thus attaining the LiDAR projection images. 
For deep multi-modal sensor fusion, the transition from different inputs to a unified modality simplifies the network structure and minimizes the fusion errors.

As shown in Fig.~\ref{fig:embedding}, there are a total of four steps in the embedding module.
The first step is resizing the camera and LiDAR matrices to $r = 384$ and $c = 384$, where $r$ is the number of rows and $c$ is the number of columns.
The second step segments the input image into non-overlapping fixed-size patches. 
The size of each patch $p$ in pixels is $16 \times 16$. 
Therefore, the dimension $\Bar{D}$ of the token representing one patch is $16 \times 16 \times 3 = 768$.
In the third step, patches are flattened into one-dimensional embedded vectors $X$ of length $\frac{r*c}{p*p} = 576$ to serve as input tokens for the transformer model.
Since transformers inherently lack the capacity to comprehend spatial and two-dimensional neighborhood structure relationships between patches, we incorporate an extra positional embedding into each patch \cite{vit}. 
The additional embedding provides the network with essential information regarding the relative spatial positions of the patches within the original image.
Sequentially, in the last step, we pass the combined patch embeddings through the Multilayer Perceptrons (MLPs) with dimensions of $E = \Bar{D} \times D$. 
$D$ indicates the network's various feature dimensions for different network parameter configurations. 
The resulting matrix $X \times E$ is the input of the transformer encoder for further learning and processing.

\subsection{Encoder}
The essence of the transformer encoder is the Multi-Head Self-Attention (MHSA) mechanism ~\cite{transformers}, which allows the network to weigh the importance of each patch relative to each other.
With the assistance of MHSA, the neural networks effectively capture global dependencies and information by computing attention scores between all pairs of patches.
Moreover, these scores are used to generate weighted sums of the patch embeddings.
The encoder output consists of embedding matrices, each corresponding to a patch in the original image.

\begin{figure*}[ht]
    \centering
    \includegraphics[width=\textwidth]{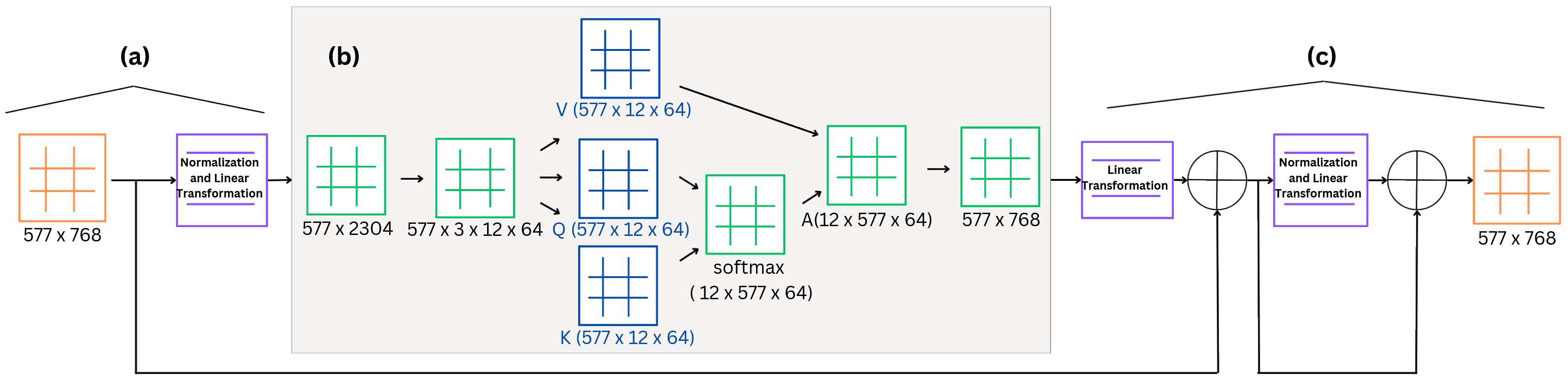}
    \caption{Encoder process. (a) The output from embedding is normalized and passed through linear layers into the multi-head attention block. (b) The matrix is split into KQV matrices, upon which SoftMax and attention operations are performed. The KQV matrices are then reshaped into a single matrix. (c) Finally, linear operations are executed, and the result is processed through the MLP block. }
    \label{fig:encoder}
\end{figure*}

Figure~\ref{fig:encoder} illustrates the detailed process of our CLFT encoder. 
The input of the encoder is the resulting matrix $X^\prime = X \times E$ from the previous embedding step (see Fig. \ref{fig:encoder}(a)). 
The matrix $X^\prime$ contains the image's patch and position embeddings, as well as the learnable class tokens. 
The dimension of the $X^\prime$ is $(576 + 1) \times 768$, which means there are 576 patch embeddings and one extra position embedding.
This approach is inspired by BERTs tokenization method, which uses similar embeddings to capture contextual information within text \cite{bert}.
The multi-head $X^\prime$ matrix is then reshaped into $577 \times 3 \times 768$, which represents a Query, Key and, Value (QKV) matrix, respectively. 
Equation~\ref{eq_multihead_attention} shows the multi-head attention $H$ calculation in this step.

\begin{equation}\label{eq_multihead_attention}
    H(Q, K, V) = \bigoplus_{i=1}^{N} h_i W^O
\end{equation}
where $\bigoplus$ means concatenation of head vectors side by side with each other, and $W^O$ is the weight matrix used to linearly transform the concatenated outputs. Each head $h_i$ is calculated individually using its own set of projection matrices as follows:
\begin{equation}\label{head_calculation}
    h_i = A(Q W^Q_i, K W^K_i, V W^V_i)
\end{equation}
where $A$ denotes the attention mechanism to the queries $(Q)$, keys $(K)$, and values $(V)$. Projection matrices $W^Q_i$, $W^K_i$, and $W^V_i$ for the $i$-th head are calculated as follows:
\begin{equation}\label{eq_projection_matrices}
\begin{split}
    W^Q_i = \mathbb{R}^{(d_m \times d_k)} \\
    W^K_i = \mathbb{R}^{(d_m \times d_k)} \\
    W^V_i = \mathbb{R}^{(d_m \times d_v)}
\end{split}
\end{equation}

The Softmax attention mechanism follows the equation \ref{eq3}:
\begin{equation}\label{eq3}
    A(Q, K, V) = softmax(\frac{QK^T}{\sqrt{d_k}})V
\end{equation}
where term $QK^T$ represents the dot product of the queries and the transposed keys, generating a similarity score between each query-key pair.
Square root of the key dimension $d_k$ prevents the dot product from becoming too large, which stabilizes the gradients during training.
The Softmax function is applied to the scaled similarity scores, converting them into attention weigths, which determine the importance of each key-value pair for the given query.
Finally, the attention weights are used to compute a weighted sum of the values $V$, producing final output of the attention mechanism for each head.

The QKV matrices are then reshaped into $N \times 577 \times 64$, where $N$ stands for the number of layers defined in CLFT configuration (as shown in Table~\ref{table:clft-variants}). At last, the metrics go though the normalization and MLP layers to be the input of CLFT decoder (Fig. \ref{fig:encoder}(c)). 

Table~\ref{table:clft-variants} outlines four potential configuration options for CLFT encoder. 
The names follow the ViT conventions.
Each configuration features predefined transformer layers and a feature dimension $D$ with fixed-size tokens.
The CLFT-Hybrid configuration distinguishes itself from the others by using a ResNet50 residual network \cite{resnet} to convert $768 \times 768$ images into $14 \times 14$ patches, then flattened into one-dimensional vectors of size 196.

\begin{table}[ht]
\vspace{0.2cm}
\caption{CLFT configuration variants.}
\label{table:clft-variants}
\centering
\begin{tabular}{|c|c|c|}
  \hline
  Type & Layers & Feature dimension D \\
  \hline
  CLFT-Base & 12 & 768 \\
  \hline
  CLFT-Large & 24 & 1024 \\
  \hline
  CLFT-Huge & 32 & 1280 \\
  \hline
  CLFT-Hybrid & 12 & 768 \\
  \hline
\end{tabular}
\end{table}

\begin{figure*}[ht]
    \centering
    \includegraphics[width=\textwidth]{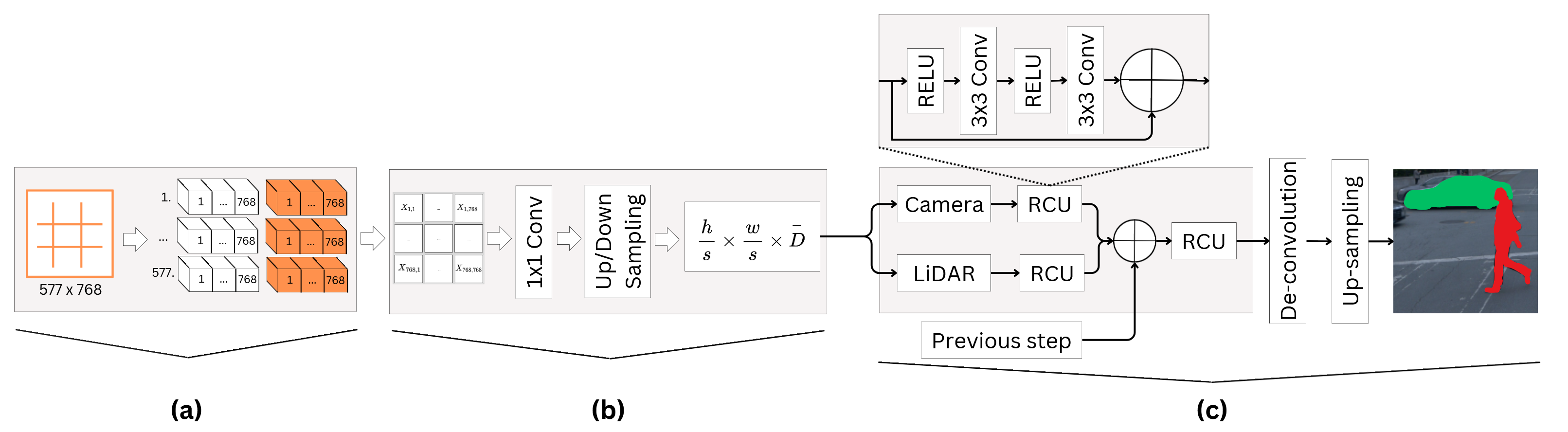}
    \caption{Decoder process. (a) The input tensor, representing data, is concatenated with classification tokens. (b) These tokens are then concatenated based on their positional information, yielding an image-like representation. Two convolution operations, along with up-sampling and down-sampling, are applied. (c) Cross-fusion is applied to combine camera and LiDAR data, progressively integrating outputs from residual computation units from previous steps. The final predicted segmentation is computed through deconvolution and up-sampling blocks.}
    \label{fig:decoder}
\end{figure*}

\subsection{Decoder}
The decoder module processes the tokens from encoder layers to progressively assemble the feature representations into a 3D matrix. 
This matrix can be visualized as an image to make predictions.
We extend the three-stage reassembly operation initially proposed in the ~\cite{dpt}, including data reading, concatenating, and resampling, with the extra stage to execute the cross-fusion of camera and LiDAR data. 

In the first stage of reassembly, shown in Fig. \ref{fig:decoder}(a), we append a special classification token to a set of $N$ tokens, potentially capturing global information.
~\cite{dpt} have evaluated three different variants of this mappings:
\begin{itemize}
    \item One that ignores the special class token and processes only the individual tokens.
    \item One that propagates information from the class token to all other tokens.
    \item One that concatenates the class token to all other tokens, then projects the combined representation through a linear layer followed by the GELU activation function to introduce non-linearity.
\end{itemize}

Figure \ref{fig:decoder}(b) shows the second stage of the decoder. 
A total amount of $N$ tokens are shaped into an image-like feature map with the aid of position tokens.
The feature map with $D$ channels is concatenated into a result $R$ = ${\frac{r}{p}}$ x ${\frac{c}{p}}$ x $D$.

Figure \ref{fig:decoder}(c) illustrates the third and last stage. The feature maps is first scaled to size $R$ = ${\frac{r}{s}}$ x ${\frac{c}{s}}$ x $\hat{D}$, where $\hat{D}$ is set as 256 in all experiments. 
Features from early layers are resampled at higher resolutions, while features from deeper layers of the transformer are resampled at lower resolutions.
The CLFT-Base variant uses layers $l = \{3,6,9,12\}$, and the CLFT-Large variant utilizes layers $l = \{5,12,18,24\}$ to extract features.
The CLFT-Hybrid variant employs ResNet layers for initial feature extraction and incorporates transformer layers $l = \{9,12\}$ for deeper feature representation.
The scaling coefficients $s$ is $\{4, 8, 16, 32\}$.

In the last cross-fusion stage, camera and LiDAR features are combined from feature maps in parallel.
Extracted feature maps are combined using the RefineNet-based feature fusion method, which employs two residual convolution units (RCUs) in a sequence.
Results from camera and LiDAR representations are summed from the previous fusion stage and passed through another RCU.
The output of the last RCU is passed to a de-convolutional layer and up-sampled to compute the predicted segmentation.

\section{\uppercase{Dataset Configuration}}
Waymo Open Datset (WOD) is designed to aid researchers in autonomous driving.
It includes data from camera and LiDAR sensors which are collected in urban and suburban environments under diverse driving conditions.
It contains labels for 4 object classes - vehicles, pedestrians, cyclists, and signs.
We have manually partitioned the dataset into four subsets: dry day, rainy day, dry night, and rainy night, and the amount of frames per subset is shown in Table \ref{table:subsets}. 
 
We use intersection over union (IoU) to evaluate the performance of the model along with values of precision and recall.
IoU computation is extended to validate multi-class semantic segmentation by assigning pixel values to void and excluding them from final validation.
We compare ground truth (Waymo label values) to the output of the CLFT model to measure the performance of our work.

\begin{table}[ht]
\vspace{0.2cm}
\caption{Frame count per subset in WOD}
\label{table:subsets}
\centering
\begin{tabular}{|c|c|c|c|}
  \hline
  Dry day & Rainy day & Dry night & Rainy night \\
  \hline
  14940 & 4520 & 1640 & 900 \\
  \hline
\end{tabular}
\end{table}

\subsection{Metrics}
We use the intersection over union (IoU) as the primary indication to evaluate the performance of our networks. 
In addition, we provide the results of precision and recall. 
The IoU is primarily used in object detection applications, in which the output is the bounding box around the object. 
We modify the ordinary IoU algorithm to fit the multi-class pixel-wise semantic object segmentation. 
Given a set of predefined semantic classes $L$ denoted by $\mathcal{L} = \{0,1,...,L-1\}$. 
Each pixel in the image can be represented as a pair $(p_{\mathcal{L}}, g_{\mathcal{L}})$, where $p_{\mathcal{L}}$ and $g_{\mathcal{L}}$ indicate the prediction and ground-truth class, respectively. 
The performance of the networks is measured by the statistics of the number of pixels that have identical classes indicated in prediction and ground truth. 
Not all pixels have a valid label, therefore ambiguous pixels that fall out of the class list are assigned as void and not counted in the evaluation. 
The IoU of each class is given by Equation \ref{eq:1}, where $\overline{\mathcal{L}}$ means the non-identical class. 

\begin{equation}
IoU_{\mathcal{L}} = \frac{\sum(p_{\mathcal{L}} g_{\mathcal{L}})}{\sum(p_{\mathcal{L}} g_{\mathcal{L}}) + \sum(p_{\mathcal{L}} g_{\overline{\mathcal{L}}}) + \sum(p_{\overline{\mathcal{L}}} g_{\mathcal{L}})}
\label{eq:1}
\end{equation}

Correspondingly, the precision and recall are obtained by Equation \ref{eq:2} and \ref{eq:3}.

\begin{equation}
Precision_{\mathcal{L}} = \frac{\sum(p_{\mathcal{L}} g_{\mathcal{L}})}{\sum(p_{\mathcal{L}} g_{\mathcal{L}}) + \sum(p_{\mathcal{L}} g_{\overline{\mathcal{L}}})}
\label{eq:2}
\end{equation}

\begin{equation}
Recall_{\mathcal{L}} = \frac{\sum(p_{\mathcal{L}} g_{\mathcal{L}})}{\sum(p_{\mathcal{L}} g_{\mathcal{L}}) +\sum(p_{\overline{\mathcal{L}}} g_{\mathcal{L}})}
\label{eq:3}
\end{equation}

\begin{table*}[ht]
\centering
\caption{Performance comparison of CLFT-Hybrid method during various weather conditions.}
\begin{tabular}{|c|c|c|c|c|c|c|c|c|c|}
\hline & \multicolumn{3}{|c|}{IoU} & \multicolumn{3}{|c|}{Precision} & \multicolumn{3}{|c|}{Recall} \\ 
\hline & Cyclist & Pedestrian & Sign & Cyclist & Pedestrian & Sign & Cyclist & Pedestrian & Sign \\ 
\hline 
\multicolumn{10}{|c|}{Dry day} \\ 
\hline 
Camera & \textbf{64.17} & 67.88 & \textbf{45.48} & \textbf{83.79} & \textbf{79.99} & \textbf{65.41} & 73.27 & \textbf{81.76} & 59.88 \\ 
LiDAR & 64.06 & \textbf{68.21} & 45.22 & 83.41 & 79.84 & 64.45 & \textbf{73.41} & 82.41 & 60.24 \\
Camera+LiDAR & 60.96 & 67.75 & 45.09 & 82.73 & 79.42 & 61.97 & 69.86 & 82.17 & \textbf{62.34} \\
\hline
\multicolumn{10}{|c|}{Rainy day} \\ 
\hline 
Camera & 70.75 & 61.98 & 35.49 & 86.19 & 80.19 & \textbf{68.98} & 79.80 & 73.18 & 42.23 \\ 
LiDAR & \textbf{73.76} & \textbf{62.84} & 37.05 & \textbf{89.53} & \textbf{80.79} & 68.02 & 80.73 & 73.89 & 44.86 \\ 
Camera+LiDAR & 72.63 & 62.50 & \textbf{37.82} & 87.27 & 79.84 & 62.30 & \textbf{81.24} & \textbf{74.22} & \textbf{49.03} \\
\hline 
\multicolumn{10}{|c|}{Dry night} \\ 
\hline 
Camera & 66.11 & 66.11 & \textbf{32.82} & 83.60 & \textbf{81.48} & 56.74 & \textbf{75.96} & 77.80 & 43.77 \\ 
LiDAR & \textbf{66.95} & \textbf{66.87} & 32.70 & \textbf{87.13} & 80.69 & \textbf{57.23} & 74.30 &\textbf{79.61} & 43.27 \\
Camera+LiDAR & 61.55 & 65.68 & 31.87 & 79.06 & 79.80 & 50.52 & 73.53 & 78.78 & \textbf{46.33} \\
\hline 
\multicolumn{10}{|c|}{Rainy night} \\ 
\hline 
Camera & 16.38 & 43.57 & \textbf{40.45} & 42.30 & 66.13 & \textbf{64.81} & 21.10 & 56.09 & 51.83 \\ 
LiDAR & 50.11 & \textbf{49.54} & 39.04 & 71.10 & 64.22 & 59.07 & 62.92 & \textbf{68.42} & 53.53 \\ 
Camera+LiDAR & \textbf{63.41} & 48.13 & 37.42 & \textbf{79.94} & \textbf{70.40} & 55.28 & \textbf{75.41} & 60.33 & \textbf{53.67} \\ 
\hline 
\end{tabular}
\label{table:hybrid_performance}
\end{table*}

\section{\uppercase{Experimental results}}
\subsection{Experimental setup}
The transformer-based networks were trained on servers equipped with Nvidia A100 80GB graphics cards.
Each training session utilized a batch size of 24, running for up to 400 epochs.
Early stopping criteria were implemented to prevent over-fitting and to ensure efficient use of computational resources.

The dataset was divided into three parts: 60\% for training, 20\% for validation, and 20\% for testing.
This distribution ensures a balanced approach, allowing the model to learn effectively, validate its performance during training, and be evaluated on unseen data to assess its generalization capabilities.

A total of nine training sessions were conducted, each with different network parameters: CLFT-Base, CLFT-Large, and CLFT-Hybrid. 
Separate training sessions were performed for LiDAR-only, camera-only, and cross-fusion of camera+LiDAR data to comprehensively evaluate the performance across different sensor configurations.

\subsection{Varying weather conditions}
We conducted an analysis of the network performance across four distinct weather conditions: dry day, rainy day, dry night, and rainy night.
The results of the CLFT-Hybrid method under these various conditions are summarized in Table \ref{table:hybrid_performance}.

In dry day conditions, the performance of the CLFT-Hybrid model using LiDAR alone (IoU: 64\% for cyclists, 68\% for pedestrians) was comparable to using camera data alone (IoU: 64\% for cyclists, 68\% for pedestrians) and slightly better than the combined data.

During rainy day conditions, LiDAR data outperformed camera data (IoU: 74\% for cyclists, 63\% for pedestrians vs. 71\% for cyclists, 62\% for pedestrians). 
This is an expected result as the camera is blurred by rain, while LiDARs are typically less affected.
Combined data was competitive, with IoU of 73\% for cyclists and 63\% for pedestrians, showing LiDAR's resilience against visual noise and low-light environments.

Under dry night conditions, LiDAR data performed better than both combined and camera data alone (IoU: 67\% for cyclists, 67\% for pedestrians vs. 66\% for cyclists and pedestrians with camera), presenting LiDAR’s advantage in low light conditions.

Under rainy night conditions, the combined LiDAR+Camera data yielded the highest performance (IoU: 63\% for cyclists, 48\% for pedestrians vs. 50\% for cyclists and 50\% for pedestrians with LiDAR alone). Cross-fusion effectively leveraged complementary information, providing depth and texture details.

\begin{table*}[ht]
\centering 
\caption{Performance metrics under dry day conditions for different CLFT configurations.} 
\begin{tabular}{|l|ccc|ccc|ccc|} 
\hline & 
\multicolumn{3}{c|}{Cyclist} & \multicolumn{3}{c|}{Pedestrian} & \multicolumn{3}{c|}{Sign} \\ 
\hline & IoU & Precision & Recall & IoU & Precision & Recall & IoU & Precision & Recall \\ 
\hline 
CLFT-Base C & 50.07 & 84.72 & 55.04 & 65.71 & 80.56 & 78.09 & 41.27 & 66.46 & 52.13 \\
CLFT-Base L & 47.01 & 84.27 & 51.53 & 64.06 & 78.60 & 77.59 & 39.76 & 63.15 & 51.78 \\
CLFT-Base C+L & 48.31 & 80.48 & 54.73 & 65.11 & 77.85 & 79.92 & 41.33 & 61.35 & 55.88 \\
\hline 
CLFT-Large C & 53.50 & 83.61 & 59.77 & 66.03 & 82.11 & 77.12 & 41.17 & 68.81 & 50.61 \\
CLFT-Large L & 53.91 & 84.53 & 59.81 & 66.31 & 80.06 & 79.43 & 41.44 & 64.49 & 53.70 \\
CLFT-Large C+L & 53.58 & \textbf{85.11} & 59.12 & 66.10 & \textbf{82.28} & 77.07 & 41.90 & \textbf{70.07} & 51.03 \\ 
\hline 
CLFT-Hybrid C & \textbf{64.17} & 83.79 & 73.27 & 67.88 & 79.99 & 81.76 & \textbf{45.48} & 65.41 & 59.88 \\
CLFT-Hybrid L & 64.06 & 83.41 & \textbf{73.41} & \textbf{68.21} & 79.84 & \textbf{82.41} & 45.22 & 64.45 & 60.24 \\
CLFT-Hybrid C+L & 60.96 & 82.73 & 69.86 & 67.75 & 79.42 & 82.17 & 45.09 & 61.97 & \textbf{62.34} \\
\hline 
\end{tabular}
\label{table:day_dry_performance} 
\end{table*}

\subsection{Varying network configurations}
The performance metrics of different CLFT configurations under dry day conditions are summarized in Table \ref{table:day_dry_performance}.
The CLFT-Base configuration showed that using either camera or LiDAR alone provides comparable results, but combining them did not yield significant improvements.
The CLFT-Large configuration benefited from higher precision, especially when combining data sources, suggesting better accuracy in identifying objects, though IoU did not significantly improve.
The CLFT-Hybrid configuration performed the best overall, particularly using either camera data alone or LiDAR data alone. 
This model effectively leverages the strengths of both data types, with the fusion of both data sources yielding high recall for signs.

\subsection{Comparison to other networks}
We compared our results to those of traditional Fully Convolutional Networks (FCN) ~\cite{clfcn} and panoptic networks as presented in ~\cite{clft}.
The CLFT-Hybrid achieved higher IoU scores (e.g., 64\% for cyclists and 68\% for pedestrians in dry day conditions) compared to typical FCN and panoptic networks, which often struggle with complex scenes and poor visibility.
Unlike FCNs and panoptic networks that rely on single modalities, the CLFT effectively combines LiDAR and camera data, enhancing performance, especially in challenging scenarios like rainy nights (IoU: 63\% for cyclists).

\section{\uppercase{Conclusion}}
\label{sec:conclusion}
In this paper, we demonstrated the effectiveness of Camera-LiDAR Fusion Transformer (CLFT) models in achieving successful object segmentation by leveraging sensor cross-fusion and the transformer's multi-attention mechanism.
The CLFT-Hybrid model showed remarkable improvements in segmentation accuracy for cyclists, pedestrians, and traffic signs.
The CLFT models maintained high performance across a variety of weather conditions, including day, rain, and night scenarios.
By combining the strengths of both LiDAR and camera data, the CLFT model effectively utilized cross-fusion to enhance overall performance.
The transformer's multi-attention mechanism enabled the CLFT models to focus on relevant features and improve object detection and segmentation accuracy.

Despite these promising results, several challenges remain.
The CLFT models exhibited variability in performance under adverse weather conditions. For instance, while LiDAR alone performed well in fair conditions, the fusion of LiDAR and camera data sometimes led to suboptimal results.
The models showed decreased performance in night and rainy conditions.
The CLFT models, especially larger configurations, require significant computational resources, which poses challenges for real-time implementation in resource-constrained environments.

Future work should focus on improving the accuracy of CLFT models in challenging environments, exploring more data fusion techniques, and integrating additional sensor modalities to further enhance overall performance.

\section*{Acknowledgment}
Part of the research has received funding from the following grants: the European Union's Horizon 2020 Research and Innovation Programme project Finest Twins (grant No. 856602) and AI-Enabled Data Lifecycles Optimization and Data Spaces Integration for Increased Efficiency and Interoperability PLIADES, grant agreement No. 101135988.

\bibliographystyle{apalike}
{\small
\bibliography{references}}

\begin{thebibliography}{}

\bibitem[Caltagirone et~al., 2018]{related-road-detection-2}
Caltagirone, L., Bellone, M., Svensson, L., and Wahde, M. (2018).
\newblock Lidar-camera fusion for road detection using fully convolutional
  neural networks.

\bibitem[Chen et~al., 2023]{3d-paper-2}
Chen, X., Zhang, T., Wang, Y., Wang, Y., and Zhao, H. (2023).
\newblock Futr3d: A unified sensor fusion framework for 3d detection.

\bibitem[Cheng et~al., 2019]{related-depth-completion-1}
Cheng, X., Wang, P., Guan, C., and Yang, R. (2019).
\newblock Cspn++: Learning context and resource aware convolutional spatial
  propagation networks for depth completion.

\bibitem[Cui et~al., 2022]{categories}
Cui, Y., Chen, R., Chu, W., Chen, L., Tian, D., Li, Y., and Cao, D. (2022).
\newblock Deep learning for image and point cloud fusion in autonomous driving:
  A review.
\newblock {\em IEEE Transactions on Intelligent Transportation Systems},
  23(2):722–739.

\bibitem[Devlin et~al., 2019]{bert}
Devlin, J., Chang, M.-W., Lee, K., and Toutanova, K. (2019).
\newblock Bert: Pre-training of deep bidirectional transformers for language
  understanding.

\bibitem[Dosovitskiy et~al., 2021]{vit}
Dosovitskiy, A., Beyer, L., Kolesnikov, A., Weissenborn, D., Zhai, X.,
  Unterthiner, T., Dehghani, M., Minderer, M., Heigold, G., Gelly, S.,
  Uszkoreit, J., and Houlsby, N. (2021).
\newblock An image is worth 16x16 words: Transformers for image recognition at
  scale.

\bibitem[Gu et~al., 2024]{clft}
Gu, J., Bellone, M., Pivoňka, T., and Sell, R. (2024).
\newblock Clft: Camera-lidar fusion transformer for semantic segmentation in
  autonomous driving.
\newblock {\em IEEE Transactions on Intelligent Vehicles}, pages 1--12.

\bibitem[Gu et~al., 2022]{clfcn}
Gu, J., Bellone, M., Sell, R., and Lind, A. (2022).
\newblock Object segmentation for autonomous driving using iseauto data.
\newblock {\em Electronics}, 11(7).

\bibitem[Gu et~al., 2018]{related-result-fusion-2}
Gu, S., Lu, T., Zhang, Y., Alvarez, J.~M., Yang, J., and Kong, H. (2018).
\newblock 3-d lidar + monocular camera: An inverse-depth-induced fusion
  framework for urban road detection.
\newblock {\em IEEE Transactions on Intelligent Vehicles}, 3(3):351--360.

\bibitem[He et~al., 2015]{resnet}
He, K., Zhang, X., Ren, S., and Sun, J. (2015).
\newblock Deep residual learning for image recognition.

\bibitem[Jaritz et~al., 2020]{related-result-fusion-1}
Jaritz, M., Vu, T.-H., de~Charette, R., Émilie Wirbel, and Pérez, P. (2020).
\newblock xmuda: Cross-modal unsupervised domain adaptation for 3d semantic
  segmentation.

\bibitem[Lai-Dang, 2024]{2d-paper-2}
Lai-Dang, Q.-V. (2024).
\newblock A survey of vision transformers in autonomous driving: Current trends
  and future directions.

\bibitem[Lee and Park, 2021]{related-road-detection-1}
Lee, J.-S. and Park, T.-H. (2021).
\newblock Fast road detection by cnn-based camera–lidar fusion and spherical
  coordinate transformation.
\newblock {\em IEEE Transactions on Intelligent Transportation Systems},
  22(9):5802--5810.

\bibitem[Li et~al., 2022]{3d-paper-3}
Li, Z., Wang, W., Li, H., Xie, E., Sima, C., Lu, T., Yu, Q., and Dai, J.
  (2022).
\newblock Bevformer: Learning bird's-eye-view representation from multi-camera
  images via spatiotemporal transformers.

\bibitem[Lin et~al., 2022]{related-depth-completion-2}
Lin, Y., Cheng, T., Zhong, Q., Zhou, W., and Yang, H. (2022).
\newblock Dynamic spatial propagation network for depth completion.

\bibitem[Ranftl et~al., 2021]{dpt}
Ranftl, R., Bochkovskiy, A., and Koltun, V. (2021).
\newblock Vision transformers for dense prediction.

\bibitem[Vaswani et~al., 2023]{transformers}
Vaswani, A., Shazeer, N., Parmar, N., Uszkoreit, J., Jones, L., Gomez, A.~N.,
  Kaiser, L., and Polosukhin, I. (2023).
\newblock Attention is all you need.

\bibitem[Wang et~al., 2021]{3d-paper-1}
Wang, Y., Guizilini, V., Zhang, T., Wang, Y., Zhao, H., and Solomon, J. (2021).
\newblock Detr3d: 3d object detection from multi-view images via 3d-to-2d
  queries.

\bibitem[Xiao and Zhu, 2023]{nlp-transformers}
Xiao, T. and Zhu, J. (2023).
\newblock Introduction to transformers: an nlp perspective.

\bibitem[Zhou and Tuzel, 2017]{related-feature-fusion-1}
Zhou, Y. and Tuzel, O. (2017).
\newblock Voxelnet: End-to-end learning for point cloud based 3d object
  detection.

\bibitem[Zhu et~al., 2024]{2d-paper-1}
Zhu, Y., Jia, X., Yang, X., and Yan, J. (2024).
\newblock Flatfusion: Delving into details of sparse transformer-based
  camera-lidar fusion for autonomous driving.

\end{thebibliography}

\end{document}